\documentclass[conference]{IEEEtran}
\usepackage{amsfonts}
\ifCLASSINFOpdf
  \usepackage[pdftex]{graphicx}
\else
\fi
\IEEEoverridecommandlockouts

\begin{document}
%
\title{A Deep Learning Approach for Facial Attribute Manipulation and Reconstruction in Surveillance and Reconnaissance}

\author{\IEEEauthorblockN{Anees Nashath Shaik}
\IEEEauthorblockA{Department of Networks \\
and Digital Media \\
Kingston University\\
London, United Kingdom \\
A.Shaik@kingston.ac.uk}
\and
\IEEEauthorblockN{Barbara Villarini}
\IEEEauthorblockA{School of Computer Science \\
and Engineering\\
University of Westminster\\
London, United Kingdom\\
b.villarini@westminster.ac.uk}
\and
\IEEEauthorblockN{Vasileios Argyriou}
\IEEEauthorblockA{Department of Networks \\
and Digital Media \\
Kingston University\\
London, United Kingdom \\
vasileios.argyriou@kingston.ac.uk}}


%


\maketitle

\IEEEpubid{\makebox[\columnwidth]{979-8-3315-1213-2/25/\$31.00 ~\copyright~2025 IEEE\hfill} \hspace{\columnsep}\makebox[\columnwidth]{ }}

\begin{abstract}
\IEEEpubidadjcol

Surveillance systems play a critical role in security and reconnaissance, but their performance is often compromised by low-quality images and videos, leading to reduced accuracy in face recognition. Additionally, existing AI-based facial analysis models suffer from biases related to skin tone variations and partially occluded faces, further limiting their effectiveness in diverse real-world scenarios. These challenges are the results of data limitations and imbalances, where available training datasets lack sufficient diversity, resulting in unfair and unreliable facial recognition performance.
To address these issues, we propose a data-driven platform that enhances surveillance capabilities by generating synthetic training data tailored to compensate for dataset biases. Our approach leverages deep learning-based facial attribute manipulation and reconstruction using autoencoders and Generative Adversarial Networks (GANs) to create diverse and high-quality facial datasets. Additionally, our system integrates an image enhancement module, improving the clarity of low-resolution or occluded faces in surveillance footage. We evaluate our approach using the CelebA dataset, demonstrating that the proposed platform enhances both training data diversity and model fairness. 
This work contributes to reducing bias in AI-based facial analysis and improving surveillance accuracy in challenging environments, leading to fairer and more reliable security applications.

\end{abstract}



%
\IEEEpeerreviewmaketitle

\section{Introduction}
Facial analysis is a fundamental component of modern surveillance and reconnaissance systems, enabling biometric identification~\cite{vv02,vv05}, security monitoring, and forensic investigations. However, these systems often struggle to perform effectively in real-world scenarios due to low-quality images and videos, which can result from poor lighting conditions, motion blur, compression artifacts, and occlusions~\cite{Schlett_2022}. Furthermore, AI-based facial recognition models exhibit performance disparities across different skin tones and partially occluded faces, limiting their reliability in diverse environment~\cite{skrishnapriya2020,vv01}. These challenges primarily originate from data limitations and biases, as many existing datasets lack sufficient representation of various demographic groups and facial conditions. As a result, surveillance models trained on such datasets may fail to generalize effectively, leading to unfair and inaccurate identification.
To overcome these limitations, we propose a data-driven platform designed to generate diverse and high-quality training data tailored to compensate for biases and dataset imbalances. Our approach leverages deep learning-based facial attribute manipulation and reconstruction using autoencoders and Generative Adversarial Networks (GANs) to synthesize realistic variations in skin tones, occlusions, and other facial features. Specifically, the proposed system includes three key components: (1) a skin tone modification model utilizing a Wasserstein GAN with Gradient Penalty (WGAN-GP) to generate diverse skin tones while preserving identity, (2) an eyeglasses removal model based on an encoder-decoder GAN architecture to improve recognition accuracy for occluded faces, and (3) a GAN-based image enhancement module incorporating style transfer techniques to refine degraded surveillance footage. We evaluated our approach using CelebA~\cite{liu2015faceattributes}, demonstrating that the proposed platform improves both the diversity of training data and the fairness of the model, leading to a significant improvement in recognition performance, particularly for underrepresented demographic groups and challenging facial conditions. Studies such as~\cite{xu2020investigating} and~\cite{menezes2021bias} highlight the impact of dataset diversity on model fairness, further supporting our approach. By addressing both data bias and image quality issues, our platform contributes to the development of fairer and more effective AI-driven surveillance systems. 
The remainder of this paper is structured as follows: Section II reviews related work on bias in facial recognition and deep learning-based image enhancement. Section III details our methodology, including dataset augmentation, GAN architectures, and image enhancement techniques. Section IV presents experimental results and performance analysis. Finally, Section V discusses key findings and future research directions, emphasizing the impact of our approach on improving fairness and effectiveness in AI-driven surveillance systems.
\section{Related Work}
The advancement of deep learning in facial analysis has led to significant improvements in surveillance and biometric identification systems. However, existing literature highlights key limitations related to data bias, low-quality imagery, and occlusion handling. This section reviews prior research on deep learning techniques for facial image enhancement, fairness in AI-driven recognition, and data augmentation strategies.

\begin{figure}[htbp]
    \centering
    \includegraphics[width=0.5\textwidth]{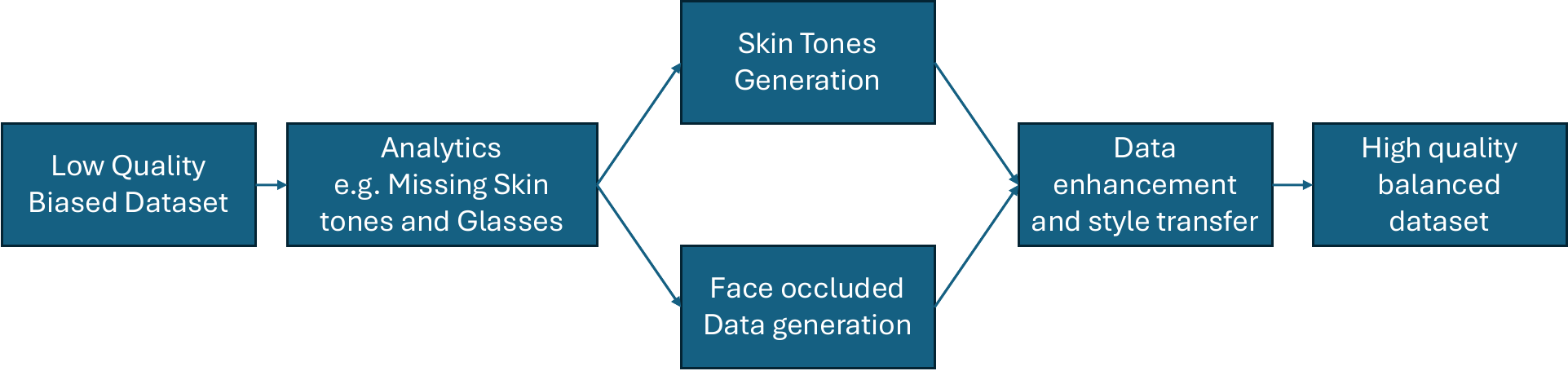}
    \caption{Overall proposed pipeline consisting of multiple stages}
    \label{fig:workflow}
\end{figure}

\subsection{Data Bias in Facial Recognition}
Bias in facial recognition models has been extensively studied, with research indicating that models often perform better on certain demographic groups due to imbalanced 
datasets~\cite{cherepanova2023deep}. 
Studies such as~\cite{terhorst2021comprehensive} demonstrate that also many non-demographic attributes strongly affect recognition performance, such as accessories, hairstyles and colors, face shapes, or facial anomalies, emphasizing the need for diverse training data to reduce disparities in facial attribute classification. Our approach builds upon these efforts by developing a data generation pipeline that expands dataset diversity and improves recognition accuracy across different skin tones and occlusion scenarios.

\subsection{Image Enhancement for Surveillance}
Low-quality surveillance footage poses a major challenge for facial recognition. Traditional methods, including super-resolution techniques~\cite{farooq2021human} and denoising algorithms~\cite{bhagwat2023empirical}, have shown improvements but struggle to generalize across varying conditions. Recent advancements in GAN-based enhancement models, such as the super-resolution GAN (SRGAN)~\cite{song2021low}, have demonstrated superior performance in refining facial details while preserving key features. Furthermore, AnimeGANv3~\cite{Liu2024dtgan} has been successfully applied for style transfer and enhancement, improving facial image clarity and definition, particularly in scenarios where traditional enhancement techniques fail. 

\subsection{Deep Learning-Based Facial Attribute Manipulation}
Generative models, particularly autoencoders and GANs, have been widely adopted for facial attribute manipulation. Works such as~\cite{senapati2023image} leverage CycleGAN and Pix2Pix architectures for image-to-image translation, enabling precise modifications while maintaining identity consistency. Additionally, facial attribute editing approaches using StyleGAN~\cite{Karras2021,vv03,vv04} provide fine-grained control over features like skin tone and occlusion handling. Recent studies have also demonstrated the effectiveness of AnimeGANv3 in generating realistic transformations, which can be applied to facial attribute modification and enhancement. Our proposed system extends these methodologies by integrating attribute manipulation with an adaptive data augmentation framework, ensuring robustness and fairness in AI-driven surveillance systems.

\section{Methodology}
The proposed approach aims to address the limitations of biased and low-quality facial datasets by implementing a structured pipeline for data enhancement and manipulation. The methodology consists of multiple stages, shown in Figure~\ref{fig:workflow}. This section outlines the technical details and deep learning techniques utilized in each stage.

The first step involves analyzing the dataset for biases, such as missing skin tones and occluded facial features (e.g., glasses, masks). This is achieved through a feature extraction model, leveraging a pre-trained ResNet to identify underrepresented attributes. The extracted features are used to inform the next stages of data augmentation and generation.

\begin{figure}[htbp]
    \centering
    \includegraphics[width=0.4\textwidth]{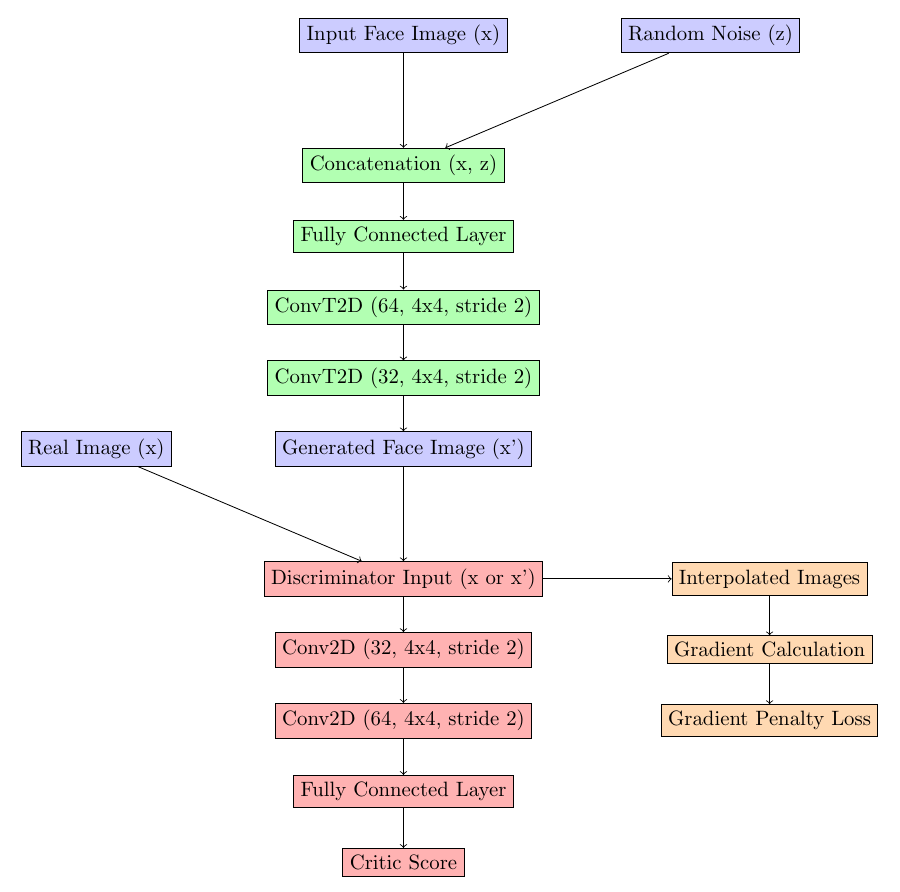}
    \caption{Skin model architecture}
    \label{fig:skin_model}
\end{figure}

\begin{figure}[htbp]
    \centering
    \includegraphics[width=0.3\textwidth]{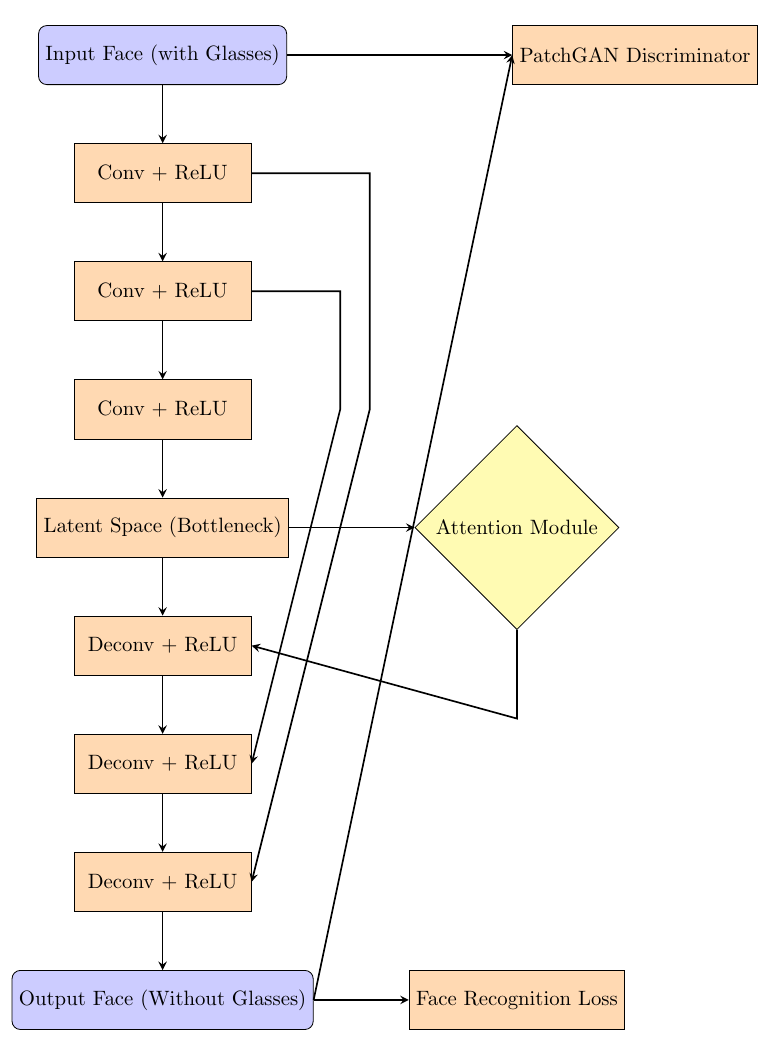}
    \caption{ERGAN architecture with encoder-decoder, skip connections, attention module, and identity loss}
    \label{fig:ERGAN}
\end{figure}

The second step aims to generate Synthetic data. The process is divided into two key components, a Skin Model and an Eyeglasses Removal Model.

\subsubsection{Skin Model Architecture}
GANs excel in image-to-image translation but often face instability issues like vanishing gradients and mode collapse. To address these challenges, the Wasserstein GAN with Gradient Penalty (WGAN-GP) \cite{gulrajani2017improved} improves training stability by enforcing the Lipschitz constraint through a gradient penalty. This study applies WGAN-GP to modify skin color in face images while preserving facial identity.
The proposed model consists of a Generator \( G \) and a Critic \( D \) (discriminator in standard GANs), which are trained adversarially. Given an input face image \( x \) and random noise \( z \), the generator produces a modified version \( G(x, z) \) with altered skin tone:

\begin{equation}
    x' = G(x, z),
\end{equation}
where \( x' \) retains structural features of \( x \) while modifying skin pigmentation.
The critic \( D \) differentiates between real images \( x \) and generated images \( x' \) by estimating the Wasserstein distance between their distributions. Instead of the traditional Jensen-Shannon divergence used in standard GANs, WGAN-GP minimizes the Wasserstein-1 distance \( W(P_r, P_g) \) between the real (\( P_r \)) and generated (\( P_g \)) distributions:

\begin{equation}
    W(P_r, P_g) = \sup_{\|f\|_L \leq 1} \mathbb{E}_{x \sim P_r} [D(x)] - \mathbb{E}_{x' \sim P_g} [D(x')],
\end{equation}

where \( f \) is a 1-Lipschitz function approximated by the critic \( D \).
To enforce the Lipschitz constraint, a gradient penalty is applied:

\begin{equation}
    \mathcal{L}_{GP} = \lambda \mathbb{E}_{\hat{x} \sim P_{\hat{x}}} \left[ (\|\nabla_{\hat{x}} D(\hat{x})\|_2 - 1)^2 \right],
\end{equation}

where \( \hat{x} \) is an interpolated sample between real and generated images:

\begin{equation}
    \hat{x} = \epsilon x + (1 - \epsilon) x', \quad \epsilon \sim \mathcal{U}(0,1).
\end{equation}

The total loss function for training the critic is then:

\begin{equation}
    \mathcal{L}_D = \mathbb{E}_{x' \sim P_g} [D(x')] - \mathbb{E}_{x \sim P_r} [D(x)] + \mathcal{L}_{GP}.
\end{equation}

The generator is trained to maximize the critic’s score:

\begin{equation}
    \mathcal{L}_G = -\mathbb{E}_{x' \sim P_g} [D(x')].
\end{equation}

The model is trained iteratively with the following steps:
\begin{enumerate}
    \item Sample real images \( x \sim P_r \) and generate fake images \( x' = G(x, z) \).
    \item Compute the critic loss \( \mathcal{L}_D \) and update \( D \).
    \item Compute the gradient penalty \( \mathcal{L}_{GP} \) and enforce the Lipschitz constraint.
    \item Update the generator by maximizing \( D(x') \).
\end{enumerate}

We use the Adam optimizer with learning rates \( \alpha_D = 1 \times 10^{-4} \) and \( \alpha_G = 1 \times 10^{-4} \), and a batch size of 64. A detailed representation of the WGan-GP model is shown in Figure~\ref{fig:skin_model}

\subsubsection{Eyeglasses Removal Model}
We develop the Eyeglasses Removal Generative Adversarial Network (ERGAN), a deep learning model designed to remove eyeglasses from facial images while preserving identity and facial structure. It follows an encoder-decoder architecture with skip connections and an attention module, ensuring high-quality, identity-preserving image generation.
ERGAN processes an input face image with glasses through a series of convolutional layers (Conv + ReLU) to extract latent feature representations. This bottleneck representation is then passed through deconvolutional layers (Deconv + ReLU) to reconstruct a glasses-free version of the face. The model incorporates skip connections, which help retain fine-grained facial details, and an attention module that focuses on the eye region to ensure precise glasses removal. Additionally, a face recognition loss function is included to maintain identity consistency between the input and output images. The realism of the generated images is further improved using a PatchGAN discriminator, which applies adversarial learning to refine the quality of the reconstructed faces. A represenatation of the overall architecture is provided in Figure~\ref{fig:ERGAN}.



\begin{figure}[htbp]
    \centering
    \includegraphics[width=0.4\textwidth]{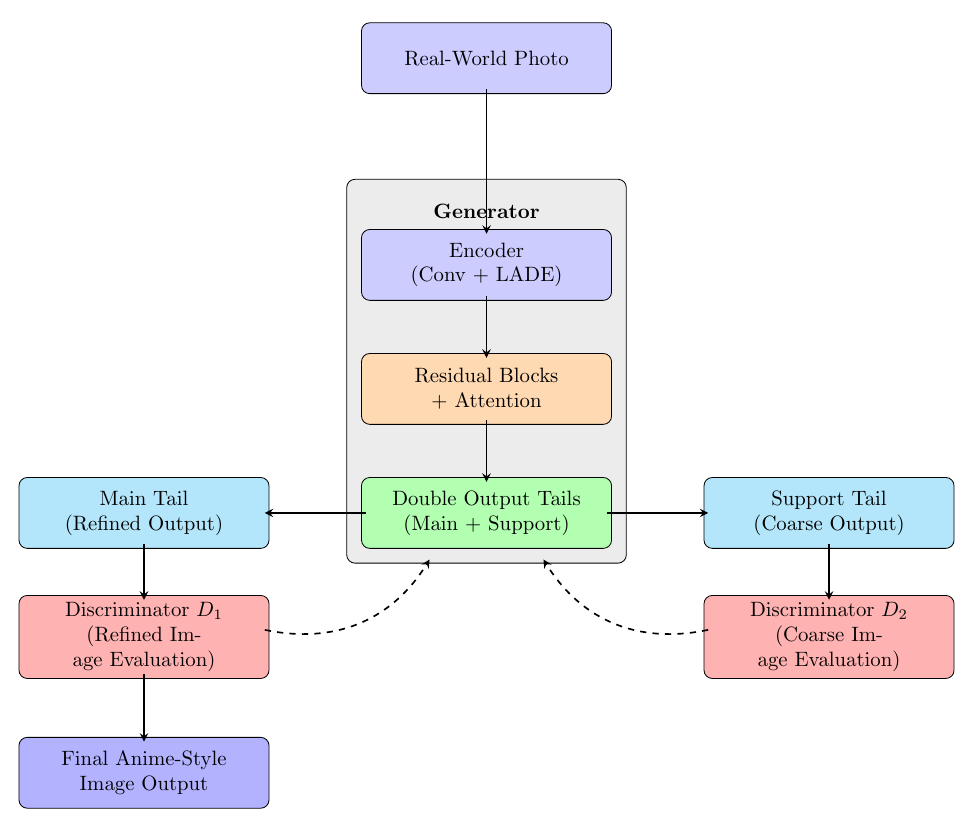}
    \caption{Overview of the AnimeGANv3 architecture, illustrating the generator's structure, dual output tails, and discriminator feedback.}
    \label{fig:Anime}
\end{figure}

\subsection{Data Enhancement and Style Transfer}
To improve the perceptual quality of surveillance footage, AnimeGANv3~\cite{Liu2024dtgan} is incorporated as a style-based enhancement framework. Unlike traditional pixel-wise restoration techniques, AnimeGANv3 employs generative adversarial networks (GANs) to refine facial structures, edges, and contours while reducing noise and artifacts. A representation of the overall architecture is illustrated in Figure~\ref{fig:Anime}.
Since AnimeGANv3 is primarily designed for artistic transformations, domain adaptation techniques are applied to align it with real-world surveillance footage. The enhancement process begins with frame extraction and resizing to 256×256 pixels for compatibility. Preprocessing techniques such as edge smoothing and superpixel segmentation are employed to reduce noise and preserve facial structures before feeding frames into the generative network.
The double-tail generator network serves as the backbone of AnimeGANv3. The support tail generates coarse-grained stylized outputs, smoothing textures and reducing unnecessary details. The main tail refines the outputs by removing artifacts and enhancing edge clarity, ensuring identity consistency across frames. A key component of the model is Linearly Adaptive Denormalization (LADE), which replaces traditional normalization techniques. Unlike instance normalization, LADE dynamically adjusts based on global statistics, preserving essential textures while preventing excessive stylization. The architecture also includes two discriminators to refine the enhancement process. The first discriminator, D1, evaluates grayscale textures, ensuring that edge structures and textures remain sharp. The second discriminator, D2, assesses overall frame clarity, preventing blurring and loss of critical details.

\subsection{High-Quality Balanced Dataset}
The final stage involves integrating the enhanced and synthetic images with the original dataset to create a well-balanced dataset. This high-quality dataset is then used for downstream tasks such as facial recognition, attribute classification, and surveillance applications.

\subsection{Dataset and Preprocessing}
The models were trained on the Large-scale CelebFaces Attributes (CelebA) dataset, a large-scale collection of over 200,000 celebrity images annotated with 40 facial attributes ~\cite{liu2015faceattributes}. CelebA offers a diverse range of facial variations, making it well-suited for learning key features such as eyeglasses. The dataset's extensive labeling allowed the model to effectively distinguish between images with and without eyeglasses, facilitating targeted feature extraction.
The dataset was split into 70\% for training, 10\% for evaluation, and 20\% for testing. To ensure consistency across training samples, all images were resized to 128×128 pixels, maintaining uniform input dimensions. Pixel values were normalized to the range [-1,1] for stable training dynamics. Data augmentation techniques, including horizontal flips and random rotations, enhanced model robustness and generalization to variations in facial orientation and expression.

\section{Results and Analysis}
The Skin Model, the Eyeglasses Removal Model and the model for Data Enhancement and Style Transfer were evaluated using two key metrics on the synthetic images generated by their respective models:  Peak Signal-to-Noise Ratio (PSNR) and Structural Similarity Index (SSIM). These metrics were chosen to assess the model's ability to reconstruct images with high fidelity.

• PSNR: Measures the quality of reconstructed images relative to the original images quantifying the level of distortions. Higher values indicate better reconstruction fidelity.

• SSIM: Evaluates the similarity between two images, commonly used in image processing and computer vision to assess perceived image quality. It compares luminance, contrast, and structural information to determine how closely the reconstructed image matches the original. SSIM values ranges from -1 to 1, where 1 indicates perfect similarity and less than 0 that the structural information is completely different.

\begin{table}
\scriptsize	
    \centering
     \caption{PSNR and its standard deviation for different categories}
    \begin{tabular}{|c|c|c|c|}
    \hline
        Metric Value & Skin & Eyeglasses & Ehnanced \\ \hline
        PSNR & 29.830 & 29.958 & 29.986 \\ \hline
        PSNR std  & 1.771 & 0.964 & 0.752 \\ \hline
    \end{tabular}
    \vspace*{3mm}
   
    \label{tab:Table1}
\end{table}

\begin{table}
\scriptsize	
    \centering
    \caption{SSIM and its standard deviation for different categories}
    \begin{tabular}{|c|c|c|c|}
    \hline
        Metric Value & Skin & Eyeglasses & Ehnanced\\ \hline
        SSIM & 0.934 & 0.385 & 0.786 \\ \hline
        SSIM std  & 0.100 & 0.195 & 0.034 \\ \hline
    \end{tabular}
    \vspace*{3mm}
    
    \label{tab:Table2}
\end{table}


The models' performance was assessed using the CelebA dataset. The numerical results are presented in Tables~\ref{tab:Table1} and ~\ref{tab:Table2}.
Images generated by the Skin Tone Modification Model achieved a PSNR of 29.830, while images from Eyeglasses Removal Model and the Image Enhancement Module yielded values of 29.958 and 29.986, respectively. These results indicate a high level of detail preservation and minimal reconstruction distortion across the different transformations. Similarly, SSIM scores revealed that the Skin Tone Model maintained strong structural consistency with an SSIM of 0.934, whereas the Eyeglasses Removal Model, which required substantial modifications to the input images, resulted in a lower SSIM of 0.385. The Image Enhancement Module achieved an SSIM of 0.786, highlighting its ability to refine facial images while preserving key features.
The qualitative analysis highlighted the model's strengths beyond the quantitative metrics. Key facial features, such as the eyes, nose, and mouth, were well-preserved without distortion. The model achieved smooth skin tone transitions and natural complexion adjustments, maintaining the face's appearance. Non-skin elements, like eyes, hair, and facial contours, were also preserved without artifacts. These results confirm that our deep learning-based approach enhances dataset diversity, improves image clarity, and reduces biases in facial recognition models, proving its robustness for real-world surveillance and reconnaissance challenges.

\section{Conclusions}
This study introduced a deep learning-based framework for improving fairness and effectiveness in AI-driven surveillance systems through facial attribute manipulation, image enhancement, and dataset augmentation. Our approach effectively addressed dataset biases, occlusions, and low-quality images, resulting in more robust and unbiased facial recognition models.
A key contribution of this work is the generation of synthetic training data to balance demographic representation, mitigating biases in AI-based facial recognition. The integration of an image enhancement module further improves recognition accuracy by refining degraded facial images. Qualitative analysis confirmed that facial features were well-preserved, ensuring structural integrity and a natural appearance.
Future work includes expanding facial attribute manipulation to incorporate expressions and aging effects, enhancing model generalization. Additionally, integrating domain adaptation techniques and real-time processing capabilities could improve applicability across various surveillance environments.
In conclusion, this work advances AI-driven surveillance by reducing biases and improving image quality, contributing to fairer and more effective facial recognition systems for security and forensic applications.

\bibliographystyle{IEEEtran}
\bibliography{refs}
%
%
%

\end{document}